\documentclass{article}

\usepackage{PRIMEarxiv}

\usepackage[utf8]{inputenc} 
\usepackage[T1]{fontenc}    
\usepackage{hyperref}       
\usepackage{url}            
\usepackage{booktabs}       
\usepackage{amsfonts}       
\usepackage{nicefrac}       
\usepackage{microtype}      
\usepackage{lipsum}
\usepackage{fancyhdr}       
\usepackage{graphicx}       
\graphicspath{{media/}}     
\usepackage{multirow}
\usepackage{subfigure}
\usepackage{threeparttable}
\usepackage{algorithm}
\usepackage{algpseudocode}
\usepackage{caption}
\usepackage{amsmath}

\title{CASOG: Conservative Actor-critic with Sm{O}oth Gradient for Skill Learning in Robot-Assisted Intervention
}

\author{
  Hao Li\\
  Institute of Automation \\
  Chinese Academy of Sciences \\
  \texttt{lihao2020@ia.ac.cn} \\
  \And
  Xiao-Hu Zhou*\\
  Institute of Automation \\
  Chinese Academy of Sciences \\
  \texttt{xiaohuu.zhou@ia.ac.cn} \\
  \And
  Xiao-Liang Xie\\
  Institute of Automation \\
  Chinese Academy of Sciences \\
  \And
  Shi-Qi Liu\\
  Institute of Automation \\
  Chinese Academy of Sciences \\
  \And
  Zhen-Qiu Feng\\
  Institute of Automation \\
  Chinese Academy of Sciences \\
  \And
  Zeng-Guang Hou*\\
  Institute of Automation \\
  Chinese Academy of Sciences \\
  \texttt{zengguang.hou@ia.ac.cn} \\
}

\begin{document}
\maketitle

\begin{abstract}
Robot-assisted intervention has shown reduced radiation exposure to physicians and improved precision in clinical trials. However, existing vascular robotic systems follow master-slave control mode and entirely rely on manual commands. This paper proposes a novel offline reinforcement learning algorithm, Conservative Actor-critic with SmOoth Gradient (CASOG), to learn manipulation skills from human demonstrations on vascular robotic systems. The proposed algorithm conservatively estimates Q-function and smooths gradients of convolution layers to deal with distribution shift and overfitting issues. Furthermore, to focus on complex manipulations, transitions with larger temporal-difference error are sampled with higher probability. Comparative experiments in a pre-clinical environment demonstrate that CASOG can deliver guidewire to the target at a success rate of 94.00\% and mean backward steps of 14.07, performing closer to humans and better than prior offline reinforcement learning methods. These results indicate that the proposed algorithm is promising to improve the autonomy of vascular robotic systems.
\end{abstract}

\section{Introduction}
\label{section: introduction}

Coronary artery disease is the most common cardiovascular diseases and kills millions every year~\cite{Mensah2019TheGB}. Percutaneous coronary intervention (PCI) is a widely used treatment for coronary artery disease. 
In PCI, physicians use X-ray fluoroscopy for guidance and deliver guidewires, catheters, and other instruments to the target vessel for treatments such as stenting and drugs. Due to X-ray fluoroscopy guidance, physicians are exposed to radiation and wear heavy lead-lined garments for radiation protection, which leads to radiation-associated hazards~\cite{Karatasakis2018RadiationassociatedLC} and orthopedic strain injuries~\cite{Klein2015OccupationalHH}. Vascular robotic systems with the master-slave control mode~\cite{Granada2011FirstinhumanEO, Guo2019ANR} have been developed to reduce the risks mentioned above. Robot-assisted intervention has shown numerous benefits in clinical trials, including X-ray exposure reduction, control precision improvement, and procedural duration decrease.

In robot-assisted intervention, instruments are shaped as flexible wires. Physicians manipulate the proximal tip of instruments outside the patient body to deliver the distal tip to the target in vessels. The relationship between manipulations and distal motion is non-linear, making instrument deliveries challenging. Learning instrument-manipulation skills and augmenting vascular robotic systems with high-level autonomy promises substantial benefits, such as shorter training periods, reduced fatigue of physicians, improved manipulation accuracy, and shorter surgery time. Due to those benefits, learning instrument-manipulation skills in robot-assisted intervention has attracted a wide range of research interests.

Human demonstrations embody rich clinical experience and are valuable for skill learning in robot-assisted intervention. Statistical models such as Gaussian Mixture Model and Hidden Markov Model are used to model the manipulations of physicians and then generate trajectories to control vascular robotic systems~\cite{rafii-tari2014hierarchical, chi2018learning}. Nevertheless, those statistical models control vascular robotic systems in an open loop and lack real-time adaptability to the vascular environment, which limits their clinical efficacy and safety. To overcome this shortcoming, behavior clone and other imitation learning methods are used to model the manipulations of physicians under different surgical scenarios~\cite{Guo2021StudyOT, Zhao2019ACP, Chi2020CollaborativeRE}. However, imitation learning methods fail to identify inferior manipulations and suffer from the unique manipulation habit of each physician.
%

Offline reinforcement learning (RL) is a learning-based control method that optimizes user-specified reward functions with previously collected static data. Compared with imitation learning, offline RL has significant advantages. Offline RL aims to maximize cumulative rewards rather than mimic offline data. Thus offline RL theoretically avoids the effects of inferior manipulations and different manipulation habits and has the potential to outperform offline data. Moreover,mathematical formalism of offline RL considers long-term effects, which is important for long-period tasks including robot-assisted intervention.
The central challenge in offline RL is distribution shift~\cite{offlineRLReview}. Distribution shift is due to the difference between the learned policy and the sampling policy for collecting offline data and could cause harmful value overestimation. Explicit regularization methods~\cite{BCQ, BRAC} directly limit the difference between the learned policy and the sampling policy to reduce distribution drift. However, these methods restrict the optimization range and may miss the optimal policy. Recent research finds that conservatively estimating the value of out-of-distribution actions can implicitly limit the learned policy close to the sampling policy with larger search range, thereby obtaining better performance~\cite{CQL, ATAC, COMBO}.
Due to these theoretical developments, offline RL has shown success in many fields such as robotic manipulation~\cite{Actionable} and auto-driving~\cite{offlineRLinAutodriving}.


For learning manipulation skills in robot-assisted intervention, offline RL faces the challenge of overfitting besides distribution shift. Since X-ray images are the primary method of intraoperative navigation, manipulation skills must use images as input. High-dimensional image input has been shown to cause overfitting and reduce the sample efficiency in both online~\cite{SAC-AE} and offline RL~\cite{DBLP:conf/l4dc/RafailovYRF21}. In online RL, data augmentation can significantly improve performance with image input~\cite{RAD, DrQ-v2}. Of all the data augmentation methods, random shift works best. Reference~\cite{A-LIX} finds that the effect of random shift is not due to the increase in data scale but to weakening overfitting caused by gradient discontinuity. The finding in the online setting provides inspiration for offline RL with image input.

The main contributions of this paper are as follows:
\begin{itemize}
  \item [1)]
  An offline RL algorithm, Conservative Actor-critic with SmOoth Gradient (CASOG), is proposed for manipulation skill learning in robot-assisted intervention.
  To the best of our knowledge, this is the first application of offline RL in this field.
  \item [2)]
  CASOG is designed to smooth gradient of convolution layers for overfitting mitigation,
  indicating potential for other offline RL problems with image input.
  \item [3)]
  Experiments show that CASOG considerably outperforms other methods and can automate guidewire deliveries with a success rate of 94.00\% and mean backward steps of 14.07. 
\end{itemize}

The paper is structured as follows: Section \ref{section: method} introduces the problem definition, offline data collection, and the proposed offline RL algorithm. Section \ref{section: results} and Section \ref{section: discussion} demonstrate and analyze the performance, respectively. Section \ref{section: conclusion} summarizes this paper.
\section{Materials and Methods}
\label{section: method}

\subsection{Problem definition}
\label{subsection: problem definition}

\begin{figure}[t!]
    \centering
    \subfigure[]{
    \includegraphics[width=0.5\linewidth]{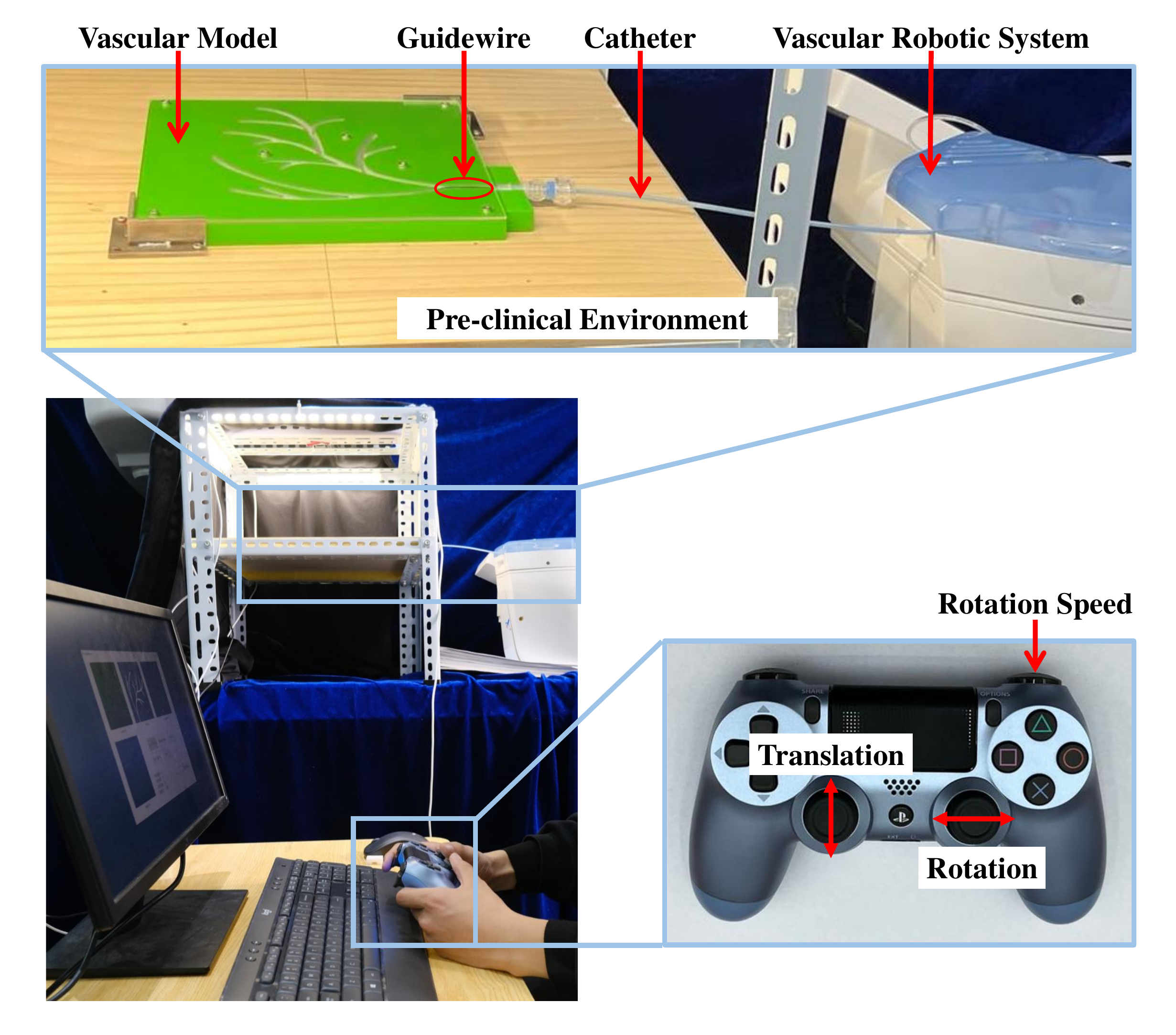}
    \label{subfig: environment}
    }
    \\
    \subfigure[]{
    \includegraphics[width=0.2\linewidth]{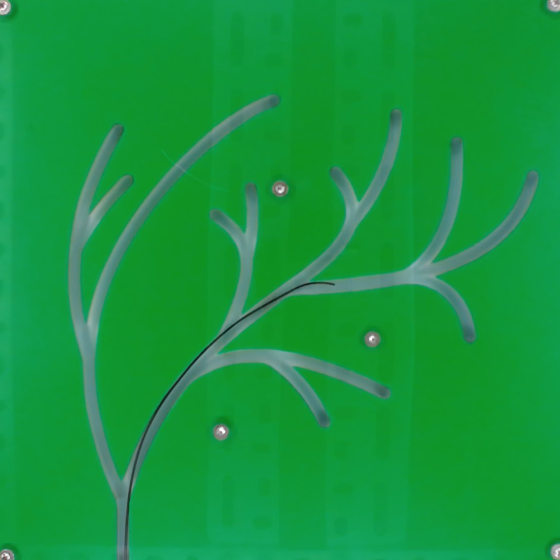}
    \label{subfig: vessel model}
    }
    \subfigure[]{
    \includegraphics[width=0.2\linewidth]{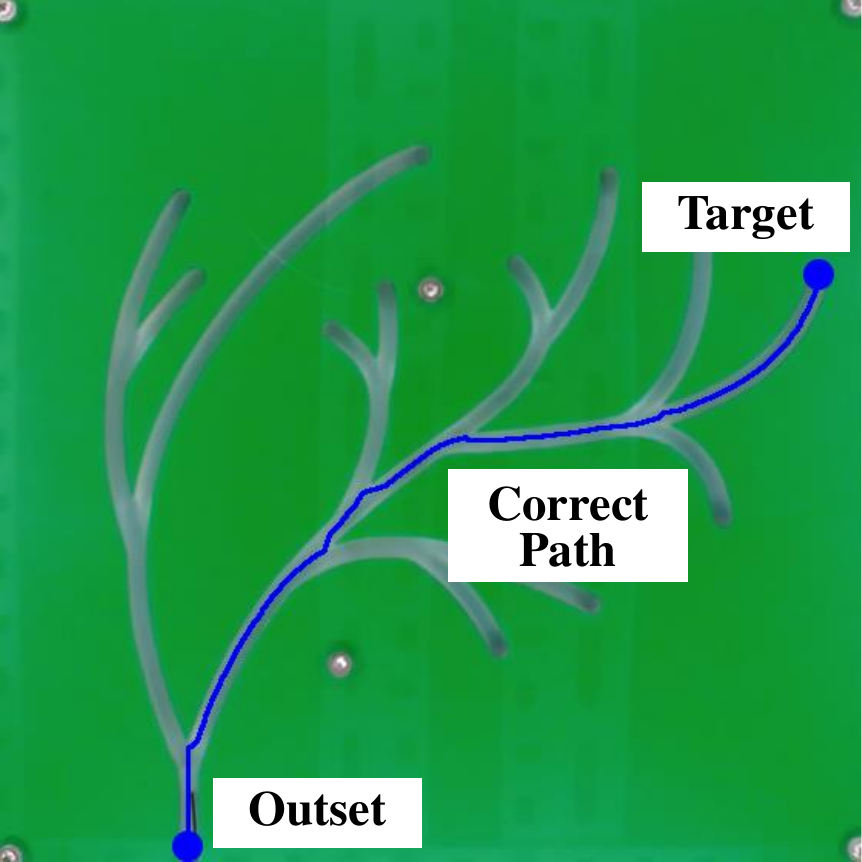}
    \label{subfig: target path}
    }
    \caption{The pre-clinical environment. (a) Offline data collection in the whole pre-clinical environment. (b) Closeup of the vascular model and guidewire. (c) The outset, target, and correct path.}
    \label{fig: environment}
\end{figure}

This paper aims to learn manipulation skills and automate guidewire deliveries in a preclinical environment similar to that in our previous work~\cite{DSAC-AE}. As shown in Fig.~\ref{subfig: environment}, the preclinical environment consists of a vascular robotic system, a Terumo RF*GA35153M guidewire, a catheter, and a 3D-printed vascular model. The guidewire is actuated by the vascular robotic system (hereinafter referred to as robot) with two degrees of freedom (translation and rotation) in the vascular model. For details of the robot, we refer readers to~\cite{Zhao2021DesignAP}. Volunteers control the robot with a gamepad for offline data collection.
Details of offline data collection are shown in Section~\ref{subsection: offline data collection}.
Fig.~\ref{subfig: vessel model} exhibits the closeup of the vascular model and guidewire.
The size of the vascular model is 15$\times$15$\times$1 cm.
The vessels are about 3$\sim$5 mm thick in the model.
The outset and the target in guidewire deliveries is shown in Fig.~\ref{subfig: target path}.

To apply offline RL, the guidewire delivery task is defined as a Partially Observable Markov Decision Process (POMDP).
A POMDP is described by the tuple $\langle\mathcal{S}, \mathcal{O}, \mathcal{A}, P, R, \gamma\rangle$, where $\mathcal{S}, \mathcal{O}, \mathcal{A}$ represent state space, action space, and observation space respectively.
$P:\mathcal{S}\times\mathcal{A}\times\mathcal{S}\mapsto\left[0,1\right]$ is the transition probability,
and $R:\mathcal{S}\times\mathcal{A}\mapsto\mathbb{R}$ is the reward function. $\gamma$ is the discount factor.
Episodes of the POMDP start in initial states where the distal tip of the guidewire is 10$\sim$15 pixels from the outset and the guidewire is rotated randomly.
At time step $t$, the robot receives an observation $o_t$ and follows policy $\pi$ to choose an action $a_t$.
\begin{figure}
    \centering
    \subfigure[]{
    \includegraphics[width=0.2\linewidth]{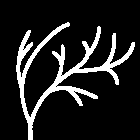}
    \label{subfig: preprocessed vessel}
    }
    \subfigure[]{
    \includegraphics[width=0.2\linewidth]{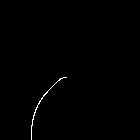}
    \label{subfig: preprocessed guidewire}
    }
    \caption{Prepossessed images as observations in POMDP. (a): The vessel. (b): The guidewire}
    \label{fig: prepossessed observation}
\end{figure}
Then the state of environment $s_t$ changes to $s_{t+1}$ according to the transition probability $P(s_{t+1}|s_t, a_t)$, and the robot gets a reward $r_t=R(s_t, a_t)$ and a new observation $o_{t+1}$.
The maximum time step is 200.
The goal is to find the optimal policy that maximizes discounted cumulative reward $\sum_{t}\mathbb{E}_{o_t,a_t}(\gamma^t r_t)$.
Q-funtion $Q(a,o)$ is the expected discounted cumulative reward with initial observation $o$ and initial action $a$, which can be regarded as an evaluation of action $a$ under observation $o$.
The details of the POMDP are as follows.

\textbf{Observation:}
The observation is the vertical view of the vascular model obtained from a fixed camera. The size of observations is 140$\times$140. 
As shown in Fig.~\ref{fig: prepossessed observation}, the observation is prepossessed into binary images of the vessel and the guidewire. 
Guidewire deliveries are successful if the distal tip of the guidewire is within 5 pixels from the target. 

\textbf{Action:}
Actions are speed commands in translation and rotation freedom.
The action space consists of ten discrete actions.
There are two sub-actions in the translation freedom, which are forward and backward with the same speed, and five sub-actions in the rotation freedom, including static, two-speed counterclockwise and clockwise rotation.
Each action is maintained for 0.5 seconds.

\textbf{Reward:}
The reward consists of the dense sub-reward and the safe sub-reward, which are designed for efficacy and safety. The dense sub-reward encourages the guidewire to approach the target along the correct path. The correct path is the shortest path from the outset to the target, as shown in Fig.~\ref{subfig: target path}, which is automatically generated by the Dijkstra algorithm. When the guidewire keeps on the correct path, the robot will get a reward equal to the decrease in distance to the target. The distance of each point to the target is calculated by the Dijkstra algorithm simultaneously with the correct path. The robot will get a penalty of size 20 if the guidewire deviates from the correct path and goes into wrong vessel branches and a bonus of size 20 if the guidewire leaves wrong vessel branches and goes back to the correct path. The robot will receive another penalty of size 50 if the guidewire exits the vascular model. The safe sub-reward limits the contact force between the robot and the guidewire. The robot will get a large penalty of size 100, and guidewire delivery will terminate if the contact force exceeds a safe threshold. The contact force is represented by the motor torque.

\subsection{Offline data collection}
\label{subsection: offline data collection}
As shown in Fig.~\ref{subfig: target path}, the guidewire delivery task is to manipulate the guidewire from the outset to the target. During data collection, volunteers observe the vertical view of the vascular model (i.e., raw images of observations in the POMDP) and choose actions using a gamepad as illustrated in Fig.~\ref{subfig: environment}. The up and down directions of the left joystick correspond to forward and backward translation, while the left and right direction of the right joystick corresponds to clockwise and counterclockwise rotation. The rotation has two possible speeds decided by the ``R2" key in the upper right corner of the gamepad. The rotation speed is high when the ``R2" key is pressed; otherwise, the rotation speed is low. According to designed actions, the rotation only takes effect when the guidewire is translated forward or backward. The offline data contains demonstrations of 7 volunteers on this task. All volunteers are right-handed men with gamepad experience but no PCI-relevant medical experience. Before data collection, each volunteer learns about the task, freely tries guidewire manipulation, and gets familiar with the manipulation mode by delivering the guidewire to each branch once. Each volunteer performs the task 6 times.
The whole offline data $\mathbb{D}$ has 42 demonstrations and 2476 time steps.
In the following, $\mathbb{E}_{\mathbb{D}}\{\cdot\}$ means the expectation that the variable obeys the distribution in the offline data $\mathbb{D}$.

\subsection{Network structure}
\label{subsection: network structure}
\begin{figure}
    \centering
    \includegraphics[width=0.5\linewidth]{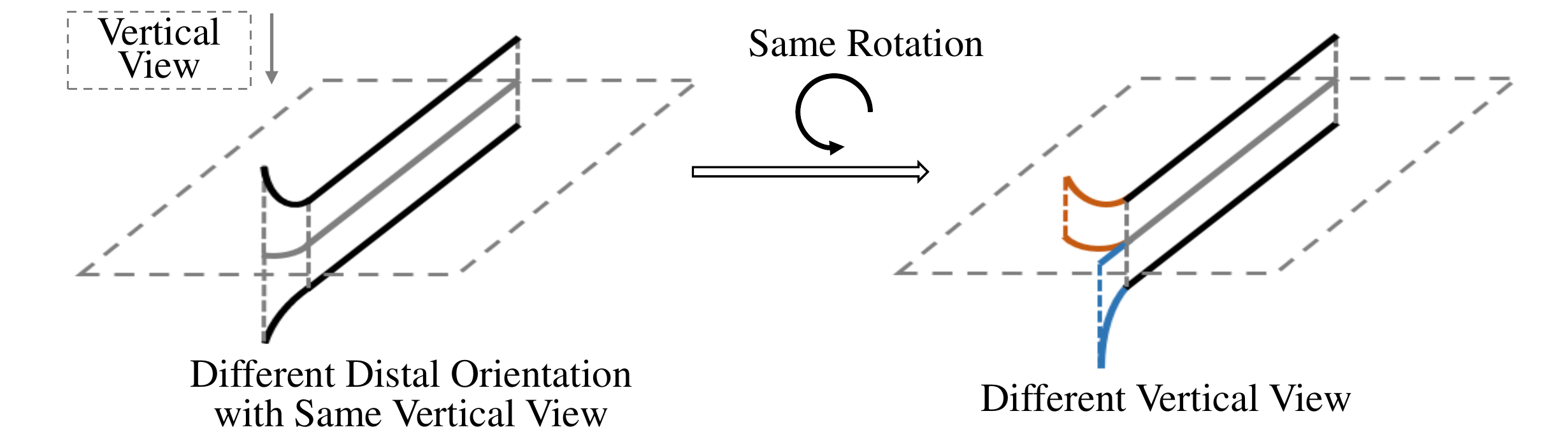}
    \caption{An example of partial observability in guidewire delivery.}
    \label{fig: partial observability}
\end{figure}

The proposed algorithm, Conservative Actor-critic with SmOoth Gradient (CASOG), follows the actor-critic framework in RL, where the policy and Q-function are parametrically estimated by neural networks and denoted by $\pi$ and $Q$. Theoretically, the optimal policy can be found when the current state is taken as the input of policy and Q-function. However, the robot can only obtain observations rather than states in the POMDP defined in Section~\ref{subsection: problem definition}. Observations contain only part of the information about states, which may affect the effectiveness of the policy and Q-function. For example,  as shown in Fig.~\ref{fig: partial observability}, the orientation of the guidewire distal tip cannot be determined entirely from the observation, and the same rotation may have different effects under the same observation. Therefore, the robot may not be able to choose the optimal action if only based on the current observation.
A common approach to deal with such partial observability is to consider the current observation as well as past actions and observations when choosing actions. Due to the small scale of offline data, the observation and action at the last time step are directly used as the additional input of the policy and Q-function rather than using recurrent neural networks. Specifically for the start time step $t=1$, $o_0=o_1$, and $a_0$ is a special action that indicates the start and cannot be selected by the policy.

Since the action space $\mathcal{A}$ is discrete, estimations of the policy and Q-function are set as $|\mathcal{A}|$-dimensional vectors denoted by $\pi(o_t|o_{t-1}, a_{t-1})$ and $Q(o_t|o_{t-1}, a_{t-1})$. Elements corresponding to action $a_t$ in $Q(o_t|o_{t-1}, a_{t-1})$ and $\pi(o_t|o_{t-1}, a_{t-1})$ are represented by $Q(o_t,a_t|o_{t-1}, a_{t-1})$ and $\pi(o_t, a_t|o_{t-1}, a_{t-1})$.

\begin{figure}
    \centering
    \includegraphics[width=0.65\linewidth]{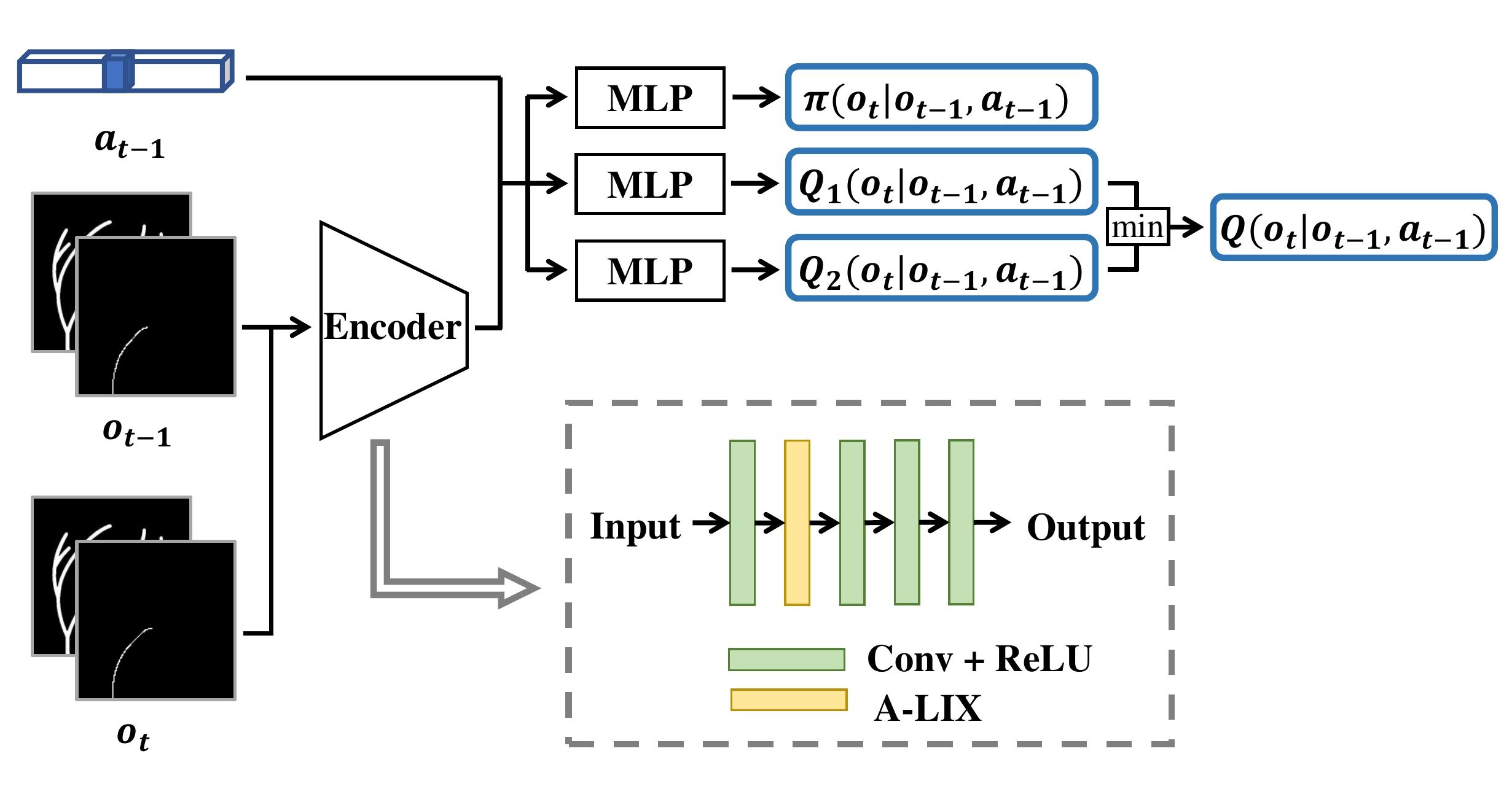}
    \caption{Network structure of CASOG}
    \label{fig: network structure}
\end{figure}

The network structure of CASOG is shown in Fig.~\ref{fig: network structure}.
The whole network consists of a shared encoder for compressing the image input and multilayer perceptions (MLP) that follow the encoder and estimate the policy and Q-function, respectively.
To reduce the error in Q-function estimation, Q-function is estimated as the minimum of two estimations $Q_1$ and $Q_2$~\cite{TD3}:
\begin{equation}
\label{equation: clipped Q}
\begin{split}
    Q(o_t,a_t|o_{t-1}, a_{t-1}) = \text{min}\left\{Q_1\left(o_t,a_t|o_{t-1}, a_{t-1}\right)\right. ,\left.Q_2\left(o_t,a_t|o_{t-1}, a_{t-1}\right)\right\}
\end{split}
\end{equation}
In the following, for writing simplicity, $o_{t-1}$ and $a_{t-1}$ are omitted. 

It is shown that smoothing gradients of convolution layers in the early stage of online RL can mitigate overfitting and improve sample efficiency~\cite{A-LIX}.
Referring to the case of online RL, the encoder in CASOG uses a special neural network layer, Adaptive Local Signal Mixing (A-LIX)~\cite{A-LIX}, to smooth the gradient adaptively.
A-LIX performs an operation similar to random shifts on feature maps obtained by convolution layers.
At training, A-LIX forwards as follows.
For input feature maps $z\in \mathbb{R}^{C\times H \times W}$ where $C$, $H$, and $W$ represent the channel number, height, and width respectively, A-LIX first samples two uniform continuous random variables $\delta_h, \delta_w \sim U[-S,S]$ representing shifts in the height and weight coordinates respectively, where $S$ denotes the shift range.
Then A-LIX performs a bilinear interpolation to obtain the output $\hat{z}$. The output element $\hat{z}\left(c,h,w\right)$, where $c$, $h$ and $w$ represent the coordinates in the channel, width and height directions, is computed as follows:
\begin{equation}
\label{equation: A-LIX}
\begin{split}
    \hat{z}\left(c,h,w\right) &= z\left(c,\lfloor\hat{h}\rfloor,\lfloor\hat{w}\rfloor\right) \left(\lceil\hat{h}\rceil-\hat{h}\right) \left(\lceil\hat{w}\rceil-\hat{w}\right) \\
    &+ z\left(c,\lfloor\hat{h}\rfloor,\lceil\hat{w}\rceil\right) \left(\lceil\hat{h}\rceil-\hat{h}\right) \left(\hat{w}-\lfloor\hat{w}\rfloor\right) \\
    &+ z\left(c,\lceil\hat{h}\rceil,\lfloor\hat{w}\rfloor\right) \left(\hat{h}-\lfloor\hat{h}\rfloor\right) \left(\lceil\hat{w}\rceil-\hat{w}\right)  \\
    &+ z\left(c,\lceil\hat{h}\rceil,\lceil\hat{w}\rceil\right) \left(\hat{h}-\lfloor\hat{h}\rfloor\right) \left(\hat{w}-\lfloor\hat{w}\rfloor\right),\\
    \hat{h} &= h + \delta_h,\\
    \hat{w} &= w + \delta_w,
\end{split}
\end{equation}
where $z(\cdot,\cdot,\cdot)$ represents an element of input feature maps. 
$\lceil\cdot\rceil$ and $\lfloor\cdot\rfloor$ represent round up and down respectively.
At testing, the output of A-LIX is equal to the input.
\begin{algorithm}[t!]
    \caption{Conservative Actor-critic with SmOoth Gradient (CASOG)}
	\label{alg: CASOG} 
	\renewcommand{\algorithmicrequire}{\textbf{Input:}}
	\renewcommand{\algorithmicensure}{\textbf{Output:}}
	\begin{algorithmic}[1]
	    \Require offline data $\{(o_{t-1}, a_{t-1}, o_t, a_t, o_{t+1}, r_{t+1})\}_t$
		\State initialize the encoder, policy, Q-function, and target Q-function;
		\State initialize transition weights;
		\Loop \Comment{imitation pretraining loop}
		\State sample transition batch using (\ref{equation: sample probability});
		\State pretrain the policy and the encoder using (\ref{equation: imitation loss});
		\State train the Q-function and the encoder using (\ref{equation: q loss});
		\State update the target Q-function;
		\EndLoop
		\Loop \Comment{offline RL loop}
		\State sample transition batch using (\ref{equation: sample probability});
		\State train the policy using (\ref{equation: pi loss});
		\State train the Q-function and the encoder using (\ref{equation: q loss});
		\State train $\alpha$ using (\ref{equation: alpha loss});
		\State train $S$ using (\ref{equation: s loss});
		\State update the target Q-function;
		\State update transition weights using (\ref{equation: transition weight});
		\EndLoop
		\Ensure the learned encoder and policy
	\end{algorithmic} 
\end{algorithm}

\subsection{Parameter update}
\label{subsection: parameter update}
Algorithm~\ref{alg: CASOG} illustrates parameter update process of CASOG, consisting of imitation pretraining loop (line 3 - line 8) and offline RL loop (line 9 - line 17). 



The main challenge in offline RL loop is how to deal with distribution shift caused by out-of-distribution (OOD) data~\cite{offlineRLReview}.
CASOG follows the idea that conservatively estimates Q-function of OOD data~\cite{CQL, ATAC, COMBO} and updates Q-function as in~\cite{ATAC} with the following loss function $J(Q)$:
\begin{equation}
\label{equation: q loss}
\begin{split}
    J(Q) =  \beta\mathbb{E}_{\mathbb{D}}\left\{\sum_{i=1,2}\left[r_t+\gamma V^{\pi}\left(o_{t+1}\right)-Q_i\left(o_t, a_t\right)\right]^2\right\} \\
    +\mathbb{E}_{\mathbb{D}}\left\{Q\left(o_t, a_t\right) - V^{\pi}\left(o_t\right)\right\},
\end{split}
\end{equation}
where $\beta$ is a hyperparemeter and value function $V^{\pi}$ is computed as follows:
\begin{equation}
    \label{equation: v function}
    \begin{split}
        V^{\pi}(o_t)=w\pi\left(o_t\right)^{\rm T}\bar{Q}\left(o_t\right) + \left(1-w\right)\pi\left(o_t\right)^{\rm T} Q\left(o_t\right).
    \end{split}
\end{equation}
$\bar{Q}$ denotes the target Q-function whose parameters are the exponentially moving average of Q-function parameters.
Hyperparameter $w$ balances $\bar{Q}$ and $Q$.
The former term in $J(Q)$ is the expectation of squared temporal-difference error (TD error) in RL theory, while the latter term can be seen as a discriminator loss that distinguishes offline data from policy $\pi$ and prevents overestimating the Q-function of OOD data.
After each optimization step of $J(Q)$, Q-function parameters are $l_2$-bounded to ensure stable learning \cite{ATAC}.
The policy $\pi$ is trained by maximizing value function $V^{\pi}$ and keeping a reasonable entropy with a Lagrangian relaxation.
The policy loss function $J(\pi)$ is as follows:
\begin{equation}
    \label{equation: pi loss}
    J(\pi) = \mathbb{E}_{\mathbb{D}}\left\{\alpha \mathcal{H}[\pi(o_t)]-V^{\pi}(o_t)\right\}, 
\end{equation}
where $\mathcal{H}(\cdot)$ stands for entropy.
$\alpha$ is a non-negative parameter updated by minimizing the following loss function:
\begin{equation}
    \label{equation: alpha loss}
    J(\alpha) = \mathbb{E}_{\mathbb{D}}\left\{\alpha\mathcal{H}[\pi(o_t)]-\alpha\bar{\mathcal{H}}\right\},
\end{equation}
where hyperparameter $\bar{\mathcal{H}}$ is the target entropy. To improve stablility, the gradient from policy is prevented to update the encoder~\cite{SAC-AE}.

In offline RL loop, shift range $S$ is automatically adjusted by a Lagrangian relaxation to keep gradient smoothness in a reasonable range.
For A-LIX output gradient $\nabla\hat{z}$, its smoothness can be measured by Modified Normalized Discontinuity Score $\widetilde{\rm ND}$~\cite{A-LIX}:
\begin{equation}
    \label{equation: ND}
    \widetilde{\rm ND}(\nabla\hat{z}) = \sum_{c=1}^{C}\sum_{h=1}^{H}\sum_{w=1}^{W}\log\left[1+\frac{D_{\nabla\hat{z}}\left(c,h,w\right)}{\nabla\hat{z}\left(c,h,w\right)^2}\right],
\end{equation}
where $C$, $H$, and $W$ represent the channel number, height, and width, respectively. $D_{\nabla\hat{z}}$ is the expected squared local discontinuity of $\nabla\hat{z}$ in any spatial direction $v$:
\begin{equation}
    \label{equation: D}
    D_{\nabla\hat{z}} = \mathbb{E}_v\left[\left(\frac{\partial \nabla\hat{z}}{\partial v}\right)^2\right].
\end{equation}
Smaller $\widetilde{\rm ND}(\nabla\hat{z})$ means that $\nabla\hat{z}$ is smoother, and $S$ is updated by minimizing the following loss function:
\begin{equation}
    \label{equation: s loss}
    J(S)=-S\times\mathbb{E}_{\mathbb{D}}\left[\widetilde{\rm ND}\left(\nabla\hat{z}\right)-\overline{\rm ND}\right],
\end{equation}
where hyperparameter $\overline{\rm ND}$ is the target of $\widetilde{\rm ND}(\nabla\hat{z})$.

Transitions with larger TD error are considered to correspond to more complex operations and are sampled for parameter updates with higher probability as in Prioritized Experience Replay (PER)~\cite{PER}. The weight of transition $i$ is exponentially moving updated after being sampled as follows:
\begin{equation}
    \label{equation: transition weight}
    f_i = \left(1-\tau\right) f_i + \tau \min\left(|\delta^i_{\rm TD}|+ \epsilon_1, \epsilon_2\right),
  \end{equation}
where $f_i$ is the weight of transition $i$. $\delta^i_{\rm TD}$ is TD error of transition $i$, and hyperparameter $\tau$ controls update speed. $\epsilon_1$ and $\epsilon_2$ are positive constants to avoid too small or too large weights.
Transition weights are initlized as $\epsilon_2$.
The probability of sampling transition $i$ is defined as follows:
\begin{equation}
    \label{equation: sample probability}
    p\left(i\right) = \frac{f_i^k}{\sum_{j=1}^{N} f_j^k},
\end{equation}
where hyperparameter $k$ controls prioritization of complex manipulations, and $N$ denotes the number of transitions.

To shorten training time, CASOG pretrains neural network parameters by imitation, which is widely used in other RL applications~\cite{AlphaGo, AlphaStar}. In imitation pretraining loop, Q-function is updated with (\ref{equation: q loss}) as in offline RL training, while the policy is updated by minimizing the imitation loss function $J_{\rm im}(\pi)$:
\begin{equation}
    \label{equation: imitation loss}
    J_{\rm im}\left(\pi\right) = \mathbb{E}_{\mathbb{D}}\left\{
    \log\left[\pi\left(o_t, a_t\right)\right]
    \right\}.
\end{equation}
The gradient from policy is used to update the encoder together with the gradient from Q-function.
$\alpha$, $S$, and transition weights are frozen in imitation pretraining loop.

\begin{table}[t!]
\caption{Value of Hyperparameters}
\label{table: hyperparameter}
\begin{center}
\begin{threeparttable}
\begin{tabular}{l l}
\hline\hline
\specialrule{0em}{2.5pt}{2.5pt}
\textbf{Hyperparameter} & \textbf{Value}\\
\specialrule{0em}{1.5pt}{1.5pt}
\hline
\specialrule{0em}{2.5pt}{2.5pt}
Hidden units of MLP & 256 \\ \specialrule{0em}{0.5pt}{0.5pt}
Number of layers of MLP & 2 \\ \specialrule{0em}{0.5pt}{0.5pt}
Channels of convolution layers & (16,32,32,32) \\ \specialrule{0em}{0.5pt}{0.5pt}
Kernel size of convolution layers & 3 \\ \specialrule{0em}{0.5pt}{0.5pt}
Stride of convolution layers & (2, 2, 1, 1) \\ \specialrule{0em}{0.5pt}{0.5pt}
Nonlinear activation & ReLU \\ \specialrule{0em}{0.5pt}{0.5pt}
Batch size & 128 \\ \specialrule{0em}{0.5pt}{0.5pt}
Optimizer & Adam \\ \specialrule{0em}{0.5pt}{0.5pt}
Learning rate of critic & 2e-4\\ \specialrule{0em}{0.5pt}{0.5pt}
Learning rate of actor & 1e-5\\ \specialrule{0em}{0.5pt}{0.5pt}
Learning rate of encoder & 2e-4\\ \specialrule{0em}{0.5pt}{0.5pt}
Discount factor $\gamma$ & 0.9 \\ \specialrule{0em}{0.5pt}{0.5pt}
Initial temperature $\alpha$ & 1.0 \\ \specialrule{0em}{0.5pt}{0.5pt}
Target entropy $\bar{\mathcal{H}}$ & 1.0 \\ \specialrule{0em}{0.5pt}{0.5pt}
Exponentially moving average of $\bar{Q}$ & 0.05 \\ \specialrule{0em}{0.5pt}{0.5pt}
$\beta$ & 2.5 \\ \specialrule{0em}{0.5pt}{0.5pt}
$w$ & 0.5 \\ \specialrule{0em}{0.5pt}{0.5pt}
$\overline{\rm ND}$ & 0.535 \\ \specialrule{0em}{0.5pt}{0.5pt}
$\left(\epsilon_1,\epsilon_2\right)$ & (1.0, 20.0) \\ \specialrule{0em}{0.5pt}{0.5pt}
$\tau$ & 0.1 \\ \specialrule{0em}{0.5pt}{0.5pt}
$k$ & 1 \\ \specialrule{0em}{0.5pt}{0.5pt}
Offline RL training steps & 50k \\ \specialrule{0em}{0.5pt}{0.5pt}
Imitation pretraining steps & 40k \\ \specialrule{0em}{0.5pt}{0.5pt}
\hline\hline
\end{tabular}
\end{threeparttable}
\label{tab1}
\end{center}
\end{table}

\section{Results}
\label{section: results}

\begin{table*}[t!]
    \renewcommand{\arraystretch}{1.3}
    \caption{Performance of CASOG and baselines}
    \centering
    \begin{threeparttable}
    \begin{tabular}{l l l l l l l}
        \hline\hline
        \specialrule{0em}{2.5pt}{2.5pt}
        \multicolumn{1}{c}{\multirow{2}{*}{Method}} & \multicolumn{1}{c}{\multirow{2}{*}{Success Rate}} & \multicolumn{2}{c}{Backward Steps}                                & \multicolumn{2}{c}{Episode Reward}                               & \multicolumn{1}{c}{\multirow{2}{*}{Time/Hours}} \\
        \specialrule{0em}{2pt}{2pt}
        \cmidrule(ll){3-4} \cmidrule(ll){5-6}
        \specialrule{0em}{1pt}{1pt}
        \multicolumn{1}{c}{}                        & \multicolumn{1}{c}{}                              & \multicolumn{1}{c}{Mean} & \multicolumn{1}{c}{STD} & \multicolumn{1}{c}{Mean} & \multicolumn{1}{c}{STD} & \multicolumn{1}{c}{} \\
        \specialrule{0em}{1.5pt}{1.5pt}
        \hline
        \specialrule{0em}{2.5pt}{2.5pt}
        CASOG (ours)                                 & \textbf{94.00\%$\pm$4.32\%}                           & 14.07 $\pm$ 2.01             & 12.51$\pm$2.14
                       & \textbf{124.49 $\pm$ 4.58}   & 16.36$\pm$12.23
                  & 1.61$\pm$0.04 \\
        \specialrule{0em}{0.5pt}{0.5pt}
        BC                                          & 56.00\%$\pm$15.75\%                                   & 39.19 $\pm$ 3.54             & 19.52$\pm$0.89
                           & 106.10 $\pm$ 11.27           & 27.95$\pm$7.73
                         & \textbackslash{} \\
        \specialrule{0em}{0.5pt}{0.5pt}
        BCQ                                         & 82.67\%$\pm$7.54\%                                    & 31.17 $\pm$ 1.35             & 21.41$\pm$0.60
                           & 122.06 $\pm$ 3.00            & 16.65$\pm$4.02
                           & 1.20$\pm$0.01 \\
        \specialrule{0em}{0.5pt}{0.5pt}
        CQL                                         & 79.33\%$\pm$2.49\%                                    & 36.43 $\pm$ 2.04             & 22.54$\pm$1.16
                        & 117.06 $\pm$ 0.60            & 28.33$\pm$4.16
                          & 1.27$\pm$0.16\\
        \specialrule{0em}{0.5pt}{0.5pt}
        Human                                       & 92.86\%                                           & \textbf{13.05}           & \textbf{10.51}                         & 123.90                    & \textbf{13.84} & \textbackslash{} \\
        \specialrule{0em}{0.5pt}{0.5pt}
        CASOG w/o PER                               & 86.00\%$\pm$2.83\%                                & 23.87$\pm$6.37           & 15.98$\pm$2.22
                         & 119.17$\pm$3.42          & 24.82$\pm$6.21
                         & 1.49$\pm$0.06 \\
        \specialrule{0em}{0.5pt}{0.5pt}
        CASOG w/o A-LIX                             & 89.33\%$\pm$5.73\%                                & 25.55$\pm$1.41           & 19.62$\pm$0.31
                         & 123.56$\pm$2.89          & 15.52$\pm$4.89
                         & 1.39$\pm$0.07 \\
        \specialrule{0em}{0.5pt}{0.5pt}
        CASOG w/o pretraining                       & 0.67\%$\pm$0.94\%                                 & \textbackslash{}         & \textbackslash{}                       & 42.03$\pm$5.47           & 35.36$\pm$7.21
                        & 1.49$\pm$0.01 \\
        \specialrule{0em}{0.5pt}{0.5pt}
        \hline\hline
    \end{tabular}
    \end{threeparttable}
    \label{table: results}
\end{table*}

Hyperparameters of CASOG are set as in Table.~\ref{table: hyperparameter}. All experiments are trained on an NVIDIA 3090 GPU and tested in 50 guidewire deliveries with 3 random seeds. The outset and the target in guidewire deliveries are set as in Fig.~\ref{subfig: target path}.

\subsection{Performance of CASOG}
\label{subsection: performance}


CASOG is compared with one imitation learning method,
Behavior Clone (BC),
and two offline RL methods,
Conservative Q-Learning (CQL)~\cite{CQL} and
Batch-Constrained deep Q-learning (BCQ)~\cite{BCQ}.
For BC, only data from successful deliveries is used and is divided into the training set and validation set with a ratio of 4:1. The model with the best performance on the validation set is selected as the training result of BC.
BCQ has an imitation loss so it is directly trained for 90k steps without imitation pretraining.
CQL has the same imitation pretraining steps and offline RL training steps as CASOG.

All methods are measured in success rate as well as mean and standard deviation of backward steps and episode reward. Success rate is calculated by dividing the number of successful guidewire deliveries by the total number. Backward steps refer to the number of backward translations during a successful guidewire delivery. As all experiments have the same target, backward steps are linearly related to the total number of steps but exclude simple advances far from bifurcations. All failed deliveries play the same role in success rate and are not taken into account by backward steps. For considering both successful and failed deliveries, episode reward is also used to measure performance of methods.

The results are shown in Table.~\ref{table: results}. CASOG significantly outperforms other methods in success rate (94.00\%) and mean backward steps (14.07), while slightly higher than BCQ and CQL and significantly higher than BC in mean episode reward (124.49). Besides, CASOG has an observably smaller standard deviation of backward steps (12.51) than BC, BCQ, and CQL. Additionally, as for the standard deviation of episode reward, CASOG performs markedly better than those of BCQ and CQL and slightly better than BCQ. The result indicates that CASOG can learn instrument manipulation skills better than prior methods and perform more stablely with random initialization of the environment. As for the computational complexity, CASOG takes more time than BCQ and CQL, but CASOG is still significantly faster than the prior online RL method that takes about 15 hours~\cite{DSAC-AE}.

The performance of offline data is represented as ``Human" in Table.~\ref{table: results} for comparison. As each volunteer only performed 6 guidewire deliveries, the performance has a large occasionality. Thus the standard deviation of offline data performance is not calculated. CASOG performs better in success rate and mean episode reward, while Human has fewer backward steps and a smaller standard deviation of backward steps and episode reward. However, the difference between CASOG and Human is not obvious in all performance metrics.

\subsection{Ablation experiments}
Ablation experiments are designed to demonstrate the effectiveness of Adaptive Local Signal Mixing (A-LIX),  Prioritized Experience Replay (PER), and pretraining in CASOG. For CASOG without pretraining, Offline RL training steps is set as 90k to keep total steps consistent with other settings. Other hyperparameters of all ablation experiments are the same as in Table.~\ref{table: hyperparameter}. 

Table.~\ref{table: results} presents results of ablation experiments.
CASOG has a higher success rate and mean episode reward as well as less mean and standard deviation of backward steps than all ablation settings. Moreover, the standard deviation of episode reward greatly increases without PER or pretraining but slightly decreases without A-LIX. Comparing results on all metrics, it can be seen that A-LIX, PER, and pretraining all play an important role in CASOG. 

All three components affect the performance of CASOG but to different degrees. CASOG without pretraining performs extremely poorly, succeeding only once in 150 deliveries across all three random seeds. At such a low success rate, backward steps make no sense and therefore are ignored. Compared with CASOG without A-LIX, CASOG without PER performs better in success rate and mean and standard deviation of episode reward but worse in the mean and standard deviation of backward steps. Moreover, CASOG without A-LIX takes less time than CASOG without PER. The result shows that pretraining has the most significant impact on performance, while A-LIX and PER have different importance on different metrics.

\section{Discussion}
\label{section: discussion}

\begin{table*}[t!]
    \renewcommand{\arraystretch}{1.3}
    \caption{Manipulation differences among volunteers}
    \centering
    \begin{threeparttable}
    \begin{tabular}{l l l l l l l l l}
        \hline\hline \specialrule{0em}{2.5pt}{2.5pt}
        Volunteer Performance     &  1     &  2     &  3     &  4     &  5     &  6    &  7     & Mean  \\
        \specialrule{0em}{1.5pt}{1.5pt} \hline \specialrule{0em}{2.5pt}{2.5pt}
        Success Rate        & 6/6   & 6/6   & 4/6   & 6/6   & 6/6   & 5/6  & 6/6   & 39/42 \\ \specialrule{0em}{0.5pt}{0.5pt}
        Mean Backward Steps           & 10.50 & 23.67 & 21.50 & 11.50 & 12.17 & 6.60 & 7.17  & 13.05 \\ \specialrule{0em}{0.5pt}{0.5pt}
        Mean Episode Reward   & 126.17 & 131.83 & 107.67 & 124.33 & 126.17 & 121.17 & 130.00  & 123.90 \\ \specialrule{0em}{0.5pt}{0.5pt}
        Mean High-speed Rotation & 9.17  & 0.50  & 1.50  & 10.67 & 21.83 & 6.00 & 0.00  & 7.10  \\ \specialrule{0em}{0.5pt}{0.5pt}
        Mean Low-speed Rotation  & 18.33 & 39.50 & 26.00 & 4.33  & 0.67  & 9.50 & 27.00 & 17.90  \\ \specialrule{0em}{0.5pt}{0.5pt}
        \hline\hline
    \end{tabular}
    \end{threeparttable}
    \label{table: manipulation difference}
\end{table*}
While Behavior Clone (BC) has been proven as a strong baseline in many offline RL problems \cite{D4RL}, BC performs poorly in our experiments. This may be because the offline data is collected from seven volunteers, and manipulations vary greatly among volunteers. Table. \ref{table: manipulation difference} shows the success rate and the mean of backward steps, episode reward, high-speed rotation, and low-speed rotation in guidewire deliveries of each volunteer. Five volunteers successfully completed all six guidewire deliveries, while the other two only succeeded in four and five guidewire deliveries, respectively. In addition, there are two volunteers with backward steps larger than 20, the backward steps of another two are less than 10, and the backward steps of the other three are between 10 and 20. This indicates that the manipulation quality varies widely among different volunteers. The low-quality manipulations in offline data would affect the performance of BC. What's more, the preference for rotational speed differs among volunteers. High-speed rotation is hardly used by three volunteers but is extremely preferred by another volunteer. Different manipulation habits may make the policy learned by BC deviate far from those of all volunteers, resulting in poorer performance than all volunteers. In a word, there are significant differences in the quality and habits of volunteers' manipulations, which may have serious negative effects on BC. However, the diversity of manipulations implies an exploration to the environment and is beneficial to offline RL. Thus, all three offline RL methods perform significantly better than BC.

As for episode reward, the standard deviation of all methods is large, but the differences among the mean are slight. The large standard deviation is due to random initialization and success criteria. The distal tip of the guidewire is 10$\sim$15 pixels from the outset, randomly rotated at the start of guidewire deliveries and within 5 pixels from the target at the end of successful guidewire deliveries. Thus, although the outset and target are fixed, the distance the guidewire shortens to the target varies in successful deliveries, resulting in a large standard deviation of episode reward. Disregarding early termination due to safety constraints, the mean episode reward depends only on the position of the distal tip at the end of delivery. Hence, successful deliveries have the same mean episode reward. Moreover, most failed deliveries are stuck in the triple bifurcation and have completed five-sixths of guidewire delivery, whose mean episode reward is only about 20 less than that of successful deliveries. Therefore, differences in mean episode reward across methods are slight.

\begin{figure}
    \centering
    \includegraphics[width=0.4\linewidth]{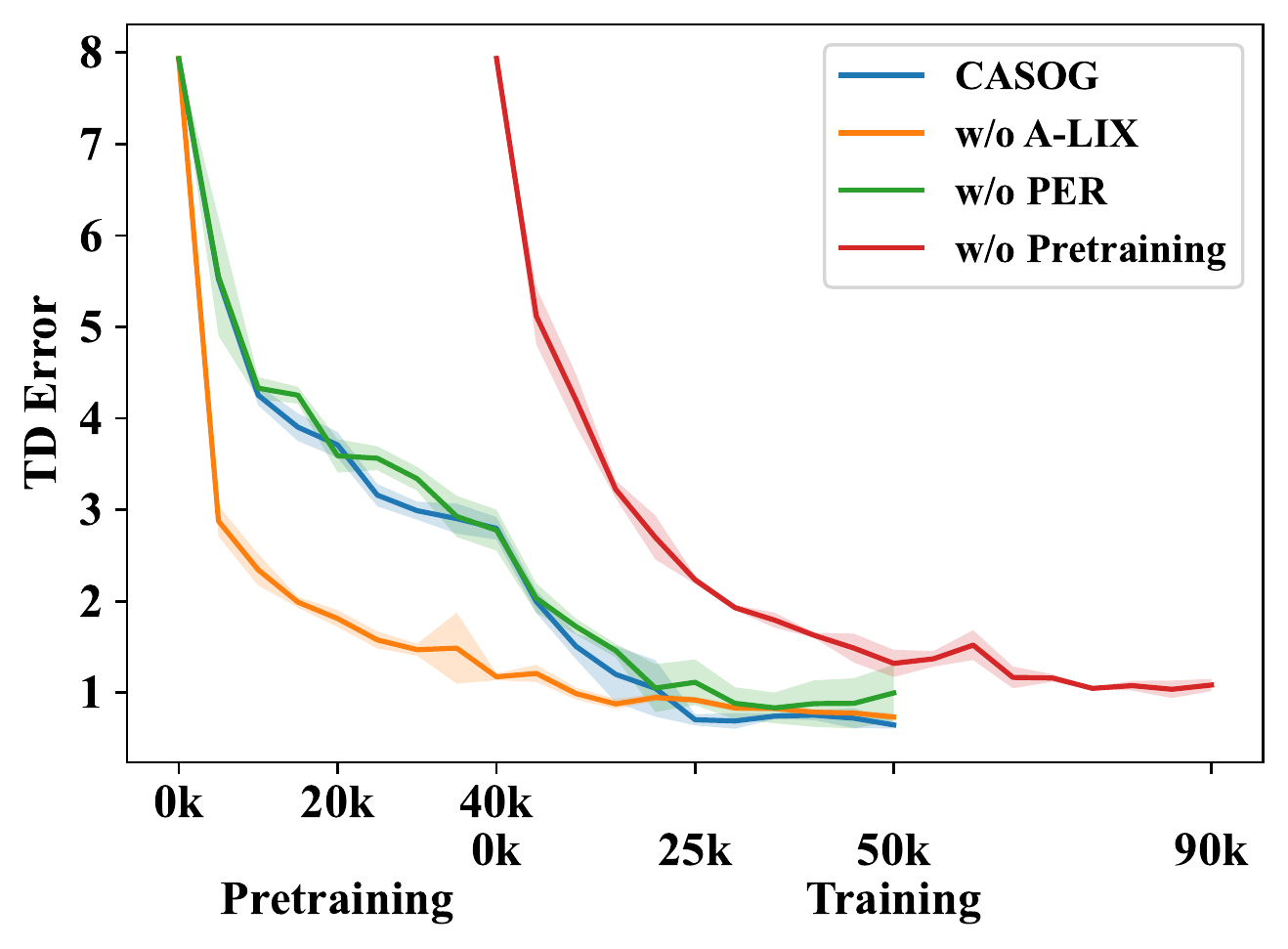}
    \caption{TD error during training. }
    \label{fig: TD error curve}
\end{figure}

Fig. \ref{fig: TD error curve} shows temporal-difference error (TD error) of CASOG and all ablation experiments. TD error is calculated with all transitions to avoid the bias induced by different sampling probabilities. In all settings, TD error is prominent in the early stage, which means Q-function cannot accurately evaluate the policy. Using Q-function to update the policy in the early stage as in (\ref{equation: pi loss}) suffers from the bias of Q-function, while imitation pretraining as in (\ref{equation: imitation loss}) avoids biased Q-function. Therefore, CASOG without pretraining performs poorly and has little success with guidewire delivery.

\begin{figure}
    \centering
    \includegraphics[width=0.4\linewidth]{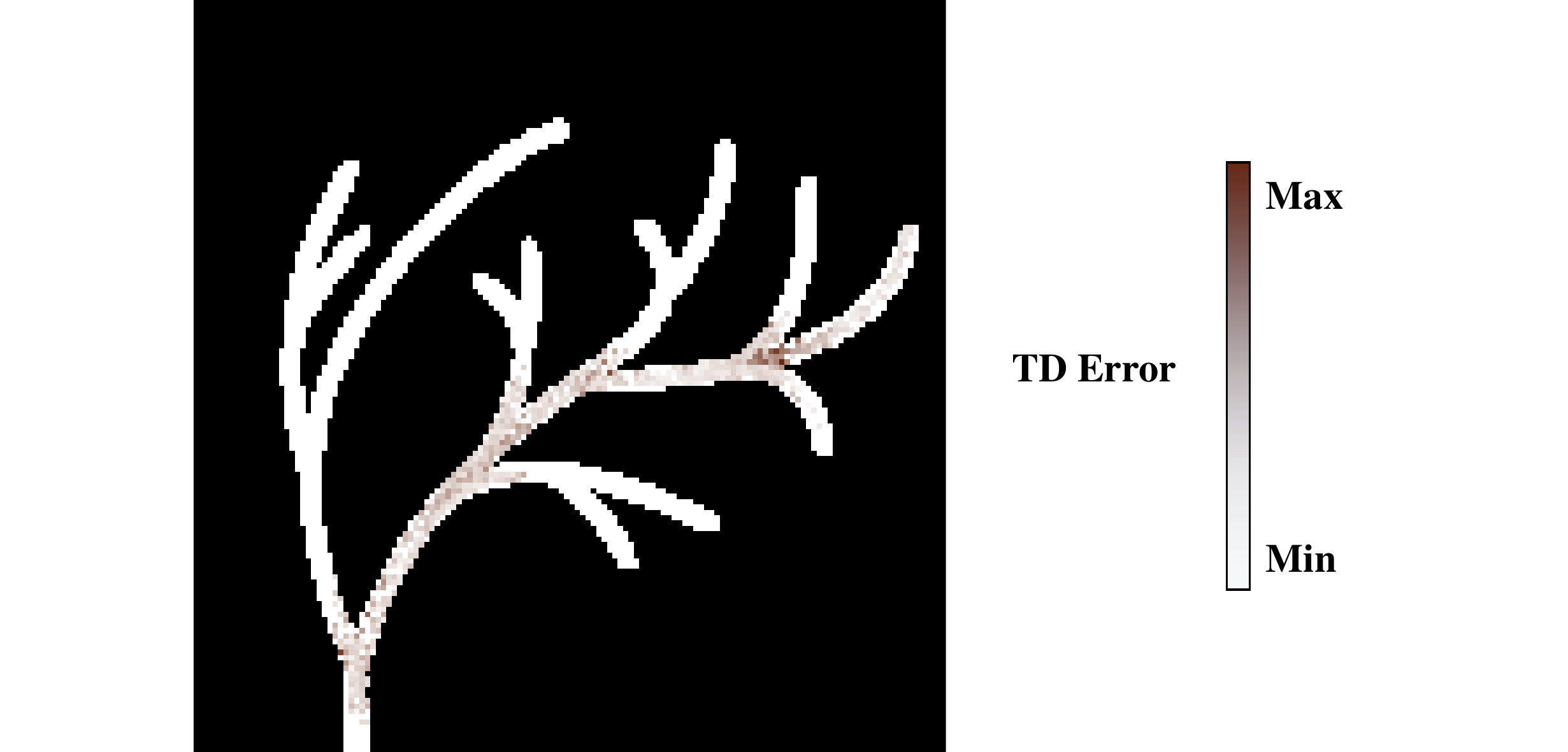}
    \caption{TD error with different distal tip positions after 5k offline RL training steps.}
    \label{fig: TD error}
\end{figure}
\begin{figure}
    \centering
    \includegraphics[width=0.4\linewidth]{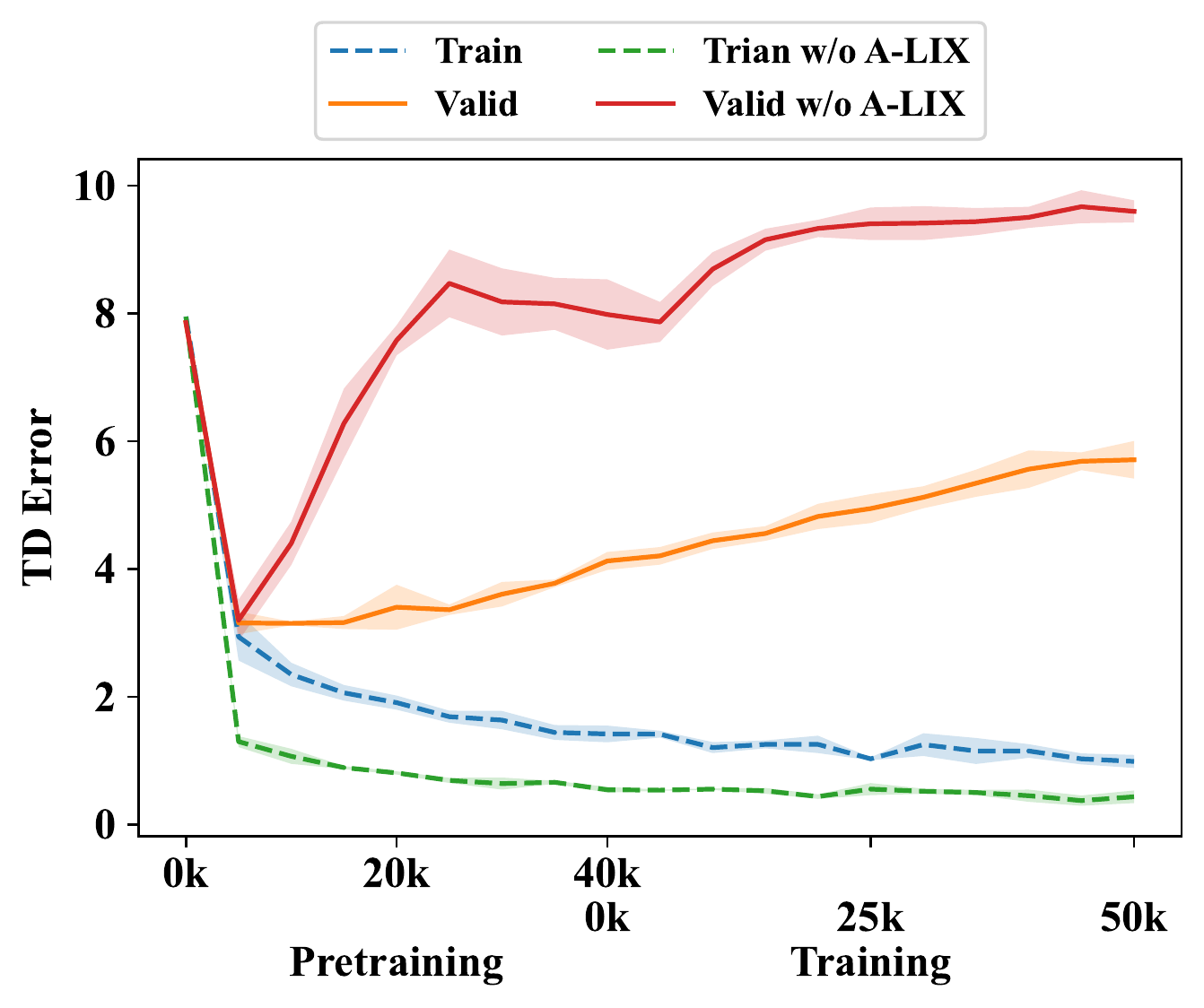}
    \caption{TD error on the training set and the valid set. }
    \label{fig: TD error curve valid}
\end{figure}

The result shows that Prioritized Experience Replay (PER) can significantly improve the performance of CASOG. To show the influence of PER, Fig. \ref{fig: TD error} demonstrates the mean TD error with different distal tip positions after 5k offline RL training steps. In the vicinity of bifurcations, especially the triple bifurcation, TD error is larger and corresponds to greater sampling probabilities. Manipulations near bifurcations are intuitively more complicated. The guidewire needs to be carefully oriented to enter the correct branch at bifurcations, while the guidewire orientation can be ignored elsewhere. Therefore, using PER, CASOG pays more attention to transitions corresponding to complex manipulations, which is beneficial for learning instrument-manipulation skills.

CASOG without Adaptive Local Signal Mixing (A-LIX) has smaller TD error during training but learns worse manipulation skills than CASOG, which may be caused by overfitting. To verify the effect of A-LIX on mitigating overfitting, CASOG and CASOG without A-LIX are trained on a training set, while TD error on a validation set is also computed. The offline data is divided into the training set and the validation set with a ratio of 4:1. TD error on the training set and the validation set is shown in Fig. \ref{fig: TD error curve valid}. Both CASOG and CASOG without A-LIX have larger TD error on the validation set than on the training set, showing that overfitting exists for both methods and may be unavoidable due to high-dimensional input and the small scale of offline data. However, CASOG without A-LIX has smaller TD error in the training set but larger TD error on the validation set than CASOG, indicating that A-LIX can effectively alleviate overfitting.
\section{Conclusion}
\label{section: conclusion}

A novel offline RL algorithm CASOG is proposed to learn instrument-manipulation skills from human demonstrations and automate guidewire delivery on vascular robotic systems. Underestimating Q-function and smoothing gradients of convolution layers in CASOG can mitigate distribution shift and overfitting. By assigning greater sampling probability to transitions with larger temporal-difference error, CASOG can pay more attention to complex manipulations near bifurcations. CASOG considerably outperforms behavior clone and prior offline RL algorithms and performs close to humans after 40k pretraining steps and 50k training steps. The subsequent work will continue to collect large-scale physician demonstrations to automate instrument deliveries in more realistic and challenging vascular models. In addition, we will try to generalize learned skills among various instruments and diverse vascular models.

\section*{Acknowledgments}
This work was supported in part by the National Natural Science Foundation of China under Grant 62003343, Grant 62222316, Grant U1913601, Grant 62073325, Grant 61720106012, Grant 62003198, Grant U20A20224, and Grant U1913210; in part by the Beijing Natural Science Foundation under Grant M22008; in part by the Youth Innovation Promotion Association of Chinese Academy of Sciences (CAS) under Grant 2020140 and in part by the Strategic Priority Research Program of CAS under Grant XDB32040000. (Corresponding author: Xiao-Hu Zhou and Zeng-Guang Hou)

\bibliographystyle{unsrt}  
\bibliography{reference}

\end{document}